%% file: main.tex
\documentclass[runningheads, a4paper, 10 pt]{llncs}  
\input{packages}

\title{Visual SLAM-based Localization and Navigation for Service Robots: The Pepper Case}

\titlerunning{Visual SLAM-based Localization and Navigation for Service Robots}

\author{Cristopher G\'omez\inst{1}\footnote{Equal contribution}  \and Mat\'ias Mattamala\inst{1\star\star} \and Tim Resink\inst{3\star\star} \and Javier Ruiz-del-Solar\inst{1,2}}

\authorrunning{Cristopher G\'omez et al.} % abbreviated author list (for running head)

\tocauthor{Cristopher G\'omez, Mat\'ias Mattamala, Tim Resink and Javier Ruiz-del-Solar}

\institute{Department of Electrical Engineering, Universidad de Chile, Chile\\
\email{\{cristopher.gomez, mmattamala, jruizd\}@ing.uchile.cl}
\\[0.1cm]
\and
Advanced Mining Technology Center, Universidad de Chile, Chile
\\[0.1cm]
\and
Delft University of Technology, The Netherlands\\
\email{p.w.resink@student.tudelft.nl}
} % end of \institute section

\begin{document}
% Main title and authors
\maketitle

% Abstract
\begin{abstract}
% We propose to use a Visual-Inertial SLAM system to expand the range of environments where Pepper can effectively localize and navigate. A modified version of ORB-SLAM is used to accomplish the localization part of the problem. The position of the robot is feed into the ROS Navigation Stack for navigation.
% \begin{itemize}
% 	\item integration of Pepper odometry \& visual SLAM
% 	\item ORB-SLAM \& pepper base model \cite{Mur-Artal2015a} \cite{Zong2017a}
% 	\item scale retrieval 
% 	\item prevention of scale drift
% 	\item separate localization \& mapping algorithm
% 	\item Map reuse \& reprojection
%\end{itemize}
We propose a Visual-SLAM based localization and navigation system for service robots. Our system is built on top of the ORB-SLAM monocular system but extended by the inclusion of wheel odometry in the estimation procedures. As a case study, the proposed system is validated using the Pepper robot, whose short-range LIDARs and RGB-D camera do not allow the robot to self-localize in large environments. The localization system is tested in navigation tasks using Pepper in two different environments: a medium-size laboratory, and a large-size hall. 
\end{abstract}

%%%%%%%%%%%%%%%%%%%%%%%%%%%%%%%%%%%%%%%%%%%%%%%%%%%%%%%%%%%%%%%%%%%%%%%%%%%%%%%%%%%%%%%%%%%%%%%%%%%%%%%%%%%%
% Introduction
\section{Introduction}
%Service robots at RoboCup @Home have been an experimental platform to research and advance towards ubiquitous robots for our society. By building their own customized robots, teams have had the chance to build the ''best`` platform for the required tasks, depending on the minimal specifications and teams' resources. However, the introduction of the new RoboCup @Home Standard Platform League in 2017 with a common, commercial platform, presented the advantage of streamlining the deployments of service robots into the real world, while offering the challenging goal of providing the standard robots with the same skills as the custom ones.

Pepper is the official robot used in the RoboCup@Home Standard Platform League. It presents several advantages for human-robot interaction such as its friendly appearance but has important limitations such as its reduced sensing and computing capabilities. In contrast to custom robots which generally rely on expensive LIDARs for metric localization and navigation, which work in both indoor and outdoor environments, Pepper has short-range LIDARs and an RGB-D camera that provide reliable localization only in small indoor rooms, being unable to provide useful information to localize the robot in large environments. This is a big deal for Pepper, which is expected to be used not only in homes, but also in public places like hospitals, shopping malls, and schools. 

To address this issue, we built upon the recent advances of visual SLAM systems to develop a visual-SLAM based self-localization solution aided by wheel odometry, which allows Pepper to self-localize and navigate in large environments. The reason to include odometry in the visual estimation procedures is to recover the metric scale (unknown in typical pure-visual schemes) and to make the visual system more robust to tracking failures. This is vital for navigation tasks that require a ``continuous'' localization hypothesis to work. The proposed solution is based on an open-source visual SLAM system, ORB-SLAM \cite{Mur-Artal2015a}, which is extended by the inclusion of the wheel's odometry in the estimation procedures.

In Section \ref{sec:visual-slam} we present a brief overview of modern SLAM systems. Then, in Section \ref{sec:preliminaries} we describe some basic notation as well as relevant characteristics of the Pepper robot. Afterwards, we present our localization and navigation approach in Section \ref{sec:proposed-system}. In Section \ref{sec:results} we present two experiments of localization and navigation with the Pepper robot in different environments. Finally, Section \ref{sec:conclusions} concludes the work with discussion and recommendations for future developments along this line.

% \begin{itemize}
% 	\item Robust localization \& navigation desired.
% 	\item Limited sensor \& computational capability Pepper.
% 	\item Depth camera sensor \& laser sensor insufficient range and resolution + IR light sensitive.
% 	\item Monocular vSLAM has scale ambiguity and ¿drift error accumulation'.
% 	\item wheel base odometry pepper erroneous, accumulating  error due to wheel slipping, uneven floors, wheel diameter problems etc. 
% 	\item Sensory integration for best of both worlds (vSLAM \& odometry) and for scale recovering
% 	\item separated remote mapping and local tracking  (single thread usage pepper).
% 	\item suited for indoor and outdoor tracking in planar camera motions
% 	\item  tightly coupled odometry SLAM in BA\cite{Mur-Artal2016a}
% \end{itemize}

%%%%%%%%%%%%%%%%%%%%%%%%%%%%%%%%%%%%%%%%%%%%%%%%%%%%%%%%%%%%%%%%%%%%%%%%%%%%%%%%%%%%%%%%%%%%%%%%%%%%%%%%%%%%
\section{Visual SLAM}
\label{sec:visual-slam}
\begin{figure}[!h]
	\centering
	\includegraphics[width=1\textwidth]{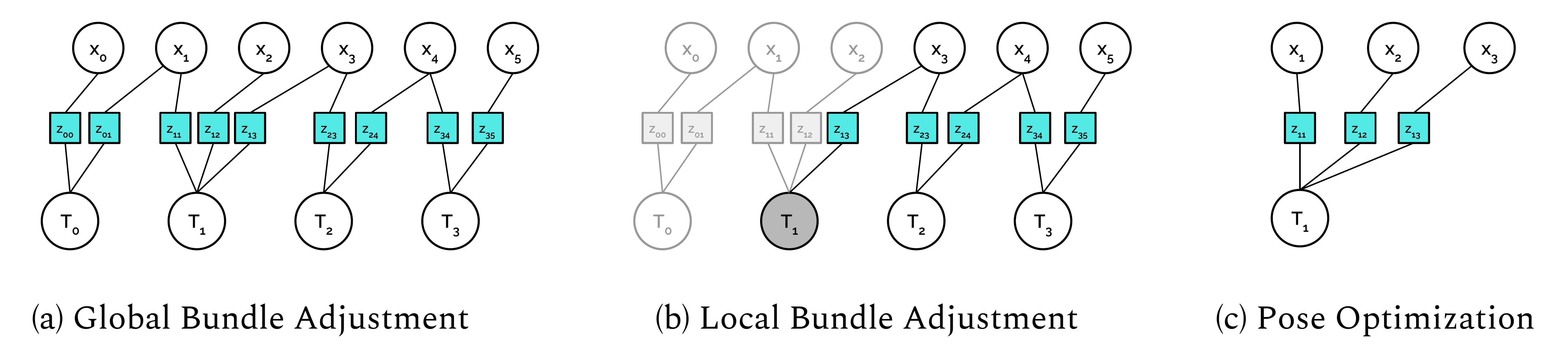}
    \vspace{-0.5cm}
	\caption{Different factor graphs related to optimization approaches in ORB-SLAM. Circles denotes variables such as map points and keyframes within a visual SLAM scheme; white are active, gray fixed. Squares denote factors or measurements.}
	\label{fig:slam}
\end{figure}

%\subsection{Localization in the RoboCup@Home}
%The standard arena in the RoboCup@Home doesn't present too much trouble for 2D LIDAR based localization. Therefore there is no need for innovation in the navigation systems. A exception to this are the challenges specifically crafted to test navigation e.g. \emph{Supermarket}, \emph{Walk \& Talk} and \emph{Help Me Carry} in which the robot has to navigate in a unknown environment.  However they can be solved with LIDAR SLAM if the test environment is not big enough to cause trouble (like in the Supermarket task). To foster the development of robots capable of working in indoor environments different than a typical house, there is a need to introduce challenges for large-scale navigation.
Visual SLAM has been a hot topic during the last years since it presents a low-cost solution for applications that require localization and mapping features such as augmented reality, virtual reality, and autonomous systems (e.g. autonomous cars, inspection drones). Being originally formulated as a filtering problem, nowadays optimization-based approaches are preferable by its superior accuracy at similar computational cost \cite{Strasdat2012}. Optimization-based approaches model the problem as a \emph{factor graph} which probabilistically relates several variables -such as poses and landmarks-, by the so-called \emph{factors}, that correspond to sensor measurements or physical constraints between the variables \cite{Cadena2016}. An example of a visual SLAM system is shown in Figure \ref{fig:slam}.
%On the contrary to its origins, the SLAM problem has changed its former filtering-based formulation into optimization-based approaches, since the latter provides the most of the accuracy at comparable computational cost \cite{Strasdat2012}. 
%Optimization-based approaches such as \emph{fixed-lag smoothers} or \emph{full smoothers} \cite{Dellaert2006} model the problem as a \emph{factor graph} which probabilistically relates several variables -such as poses and landmarks-, by the so-called \emph{factors}, that correspond to sensor measurements or physical constraints between the variables. An example of this formulation for a visual SLAM system is shown in Figure \ref{fig:slam}.

The factor graph can be formulated as a non-linear least squares problem \cite{Cadena2016} that aims to find the states $\notatManifold{\optimum{X}} = {\notationManifold{X}{1}, ..., \notatManifold{X}{m}}$ that minimize the error between the measurements $\notationManifold{z}{i}$, and an \emph{observational model} $h_{i}(\notatManifold{X}_i)$ that ''predicts`` the expected measurement given the state $\notatManifold{X}_i$ \footnote{The operator $\boxminus$ generalizes the concept of subtraction for non-Euclidean spaces. Please refer to Hertzberg \etal \cite{Hertzberg2013} for a complete treatment.}: 
\begin{equation} \label{eq:slam-nls}
\notatManifold{\optimum{X}} = \argmin_\notatManifold{X} = {\displaystyle \sum_{i=1}^{m}\ \mahalanobisNorm{ h_{i}(\notatManifold{X}_i) \boxminus \notationManifold{z}{i}}{\notationMatrix{\Omega}{i}} }
\end{equation}
The same formulation holds for the visual case, where the states correspond to selected camera poses of the trajectory -keyframes- and also the map representation -3D points, surfels, voxels, etc-, and the measurements are reprojections of the map into the image plane.

Regarding some actual systems, different solutions have been developed for monocular and stereo/depth sensors. We are concerned about monocular solutions since cameras are ubiquitous in current robots while being ``cheap'' sensors in comparison to the other two; they also can work in both indoor and outdoor environments.
Monocular visual SLAM systems are either \emph{feature-based} that utilize just some features in the image, such as ORB-SLAM \cite{Mur-Artal2015a}, or direct methods that exploit the complete information from every image as in LSD-SLAM \cite{Engel2014}.

The main issue with monocular systems is that they require a moving camera in order to estimate the depth of the scene, as well as depending on an unknown scale factor that maps the estimated states to physically consistent dimensions. The typical approach to solve the problem relies on the usage of different sources of information that provides the scale, such as inertial measurements units (IMU); however, this increases the computational requirements of the estimation problem, since the number of states increases \cite{Mur-Artal2016a}.

The utilization of visual localization systems in the RoboCup@Home has been disregarded since most of the custom robots could afford accurate but expensive LIDAR systems \cite{Iocchi2015} \cite{Cheng2015}, which provide a simpler solution. Nevertheless, since the range of Pepper's LIDARs and depth camera are defined by the manufacturer, and the RoboCup@Home SSPL (Social Standard Platform League) forbids the use of additional sensors, it is unfeasible for the robot neither localize nor navigate in large environments. For this reason, we propose a visual approach for the localization problem based on an open-source visual SLAM system, ORB-SLAM \cite{Mur-Artal2015a}, and we present a strategy to solve visual SLAM issues (mainly the lack of a metric scale) by aiding the system with wheel odometry measurements.

%%%%%%%%%%%%%%%%%%%%%%%%%%%%%%%%%%%%%%%%%%%%%%%%%%%%%%%%%%%%%%%%%%%%%%%%%%%%%%%%%%%%%%%%%%%%%%%%%%%%%%%%%%%%
\section{Platform, coordinate systems and notation}
\label{sec:preliminaries}

\subsection{Notation}
To prevent confusion in notation, we follow the conventions of Paul Furgale\footnote{\url{http://paulfurgale.info/news/2014/6/9/representing-robot-pose-the-good-
the-bad-and-the-ugly}}:
\begin{itemize}
	\item Coordinate frame A is notated as $ \notatFrame{A}$.
	\item Homogeneous transformation matrix $ \notationMatrixFrame{T}{O}{WC} \in \SEthree$ represents the pose of the camera frame $ \notatFrame{C}$ with respect to the world frame, $ \notatFrame{W}$, seen from frame  $ \notatFrame{O}$. A vector expressed in world frame W, $\notationVectorFrame{v}{W}{}$ can be hereby transformed to the camera frame C by the rotation matrix $\notationMatrix{R}{WC} \in \SOthree$, as $\notationVectorFrame{v}{C}{} = \notationMatrix{R}{WC} \ \notationVectorFrame{v}{W}{}$ 
	\item  The homogeneous transformation matrix $ \notationMatrixFrame{T}{C}{WC}$ will be abbreviated to $ \notationMatrix{T}{WC}$ for reader convenience unless otherwise indicated.
	%\item   $ \notationMatrixFrame{T}{O}{WC}$ will refer to the pose obtained from the odometry, whereas $ \notationMatrix{T}{WC}$ will refer to the ORB-SLAM's pose.
\end{itemize}

\subsection{Pepper robot}
Pepper is a wheeled humanoid platform. It has a mobile omnidirectional base and 20 degrees of freedom, including an actuated pelvis and neck. It has two Omnivision OV5640 cameras, located in the forehead and the mouth, in addition to an RGB-D sensor in the eyes. Additionally, the base has three LIDARs and an IMU. In order to access the sensors and control the robot, Softbank provides an API to its middleware, NAOQi. 

Since we base our system in the ROS framework, we access sensing and perform control through the \texttt{naoqi\_driver} ROS package. We principally use the images from the forehead camera at a 640x480 pixels resolution, as well as the internally computed odometry measurements; the algorithmic details about the latter are unknown to the user.

\begin{figure}[!t]
	\centering
	\includegraphics[width=0.4\textwidth]{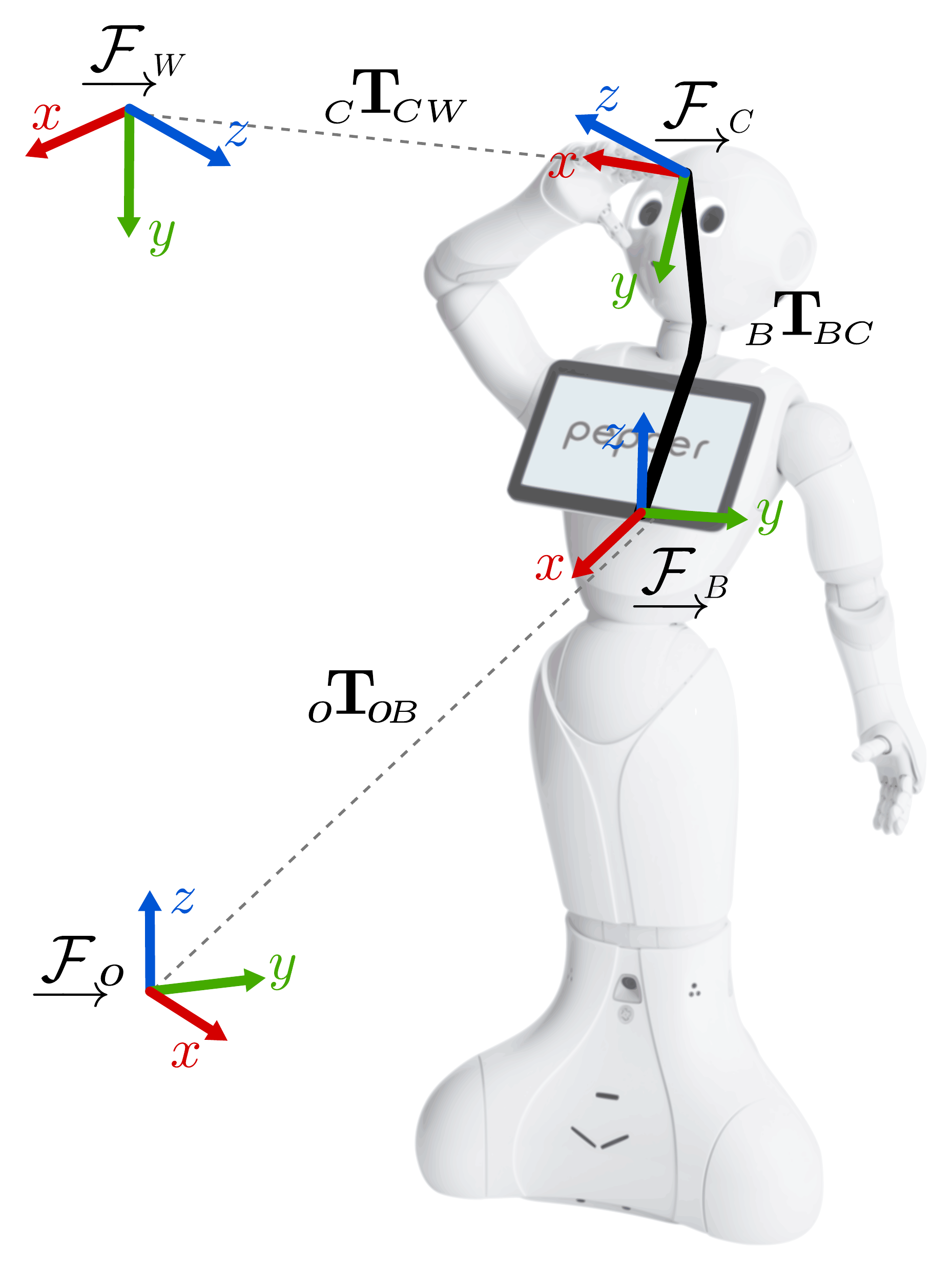}
	\caption{Coordinate frames used in this work, we follow the classic conventions with X red, Y green and Z blue. $\notatFrame{C}$ denotes the camera frame, $\notatFrame{O}$ the odometry frame and $\notatFrame{B}$ the body's. Pepper picture is based on Philippe Dureuil's.}
	\label{fig:reference-frames}
\end{figure}

We considered two main reference frames for this work (Figure \ref{fig:reference-frames}): On the one hand, the odometry frame denoted by $\notatFrame{O}$ describes the pose of the robot relative to the initial pose, as defined in ROS REP 105 \footnote{\url{http://www.ros.org/reps/rep-0105.html}}. We use this frame to describe the pose of the robot's torso (body) computed by the internal wheel odometry, denoted by $\notationMatrixFrame{T}{O}{OB}$.  On the other, ORB-SLAM has its own reference frame (world) $ \notatFrame{W}$ that depends on the initialization of the system, hence it may change every time the system is reset. The estimate provided by ORB-SLAM corresponds to the world position with respect to the camera pose,  $\notationMatrixFrame{T}{C}{CW}$.

\section{Localization and navigation system}
\label{sec:proposed-system}
\begin{figure}[t]
	\centering
	\includegraphics[width=1.0\textwidth]{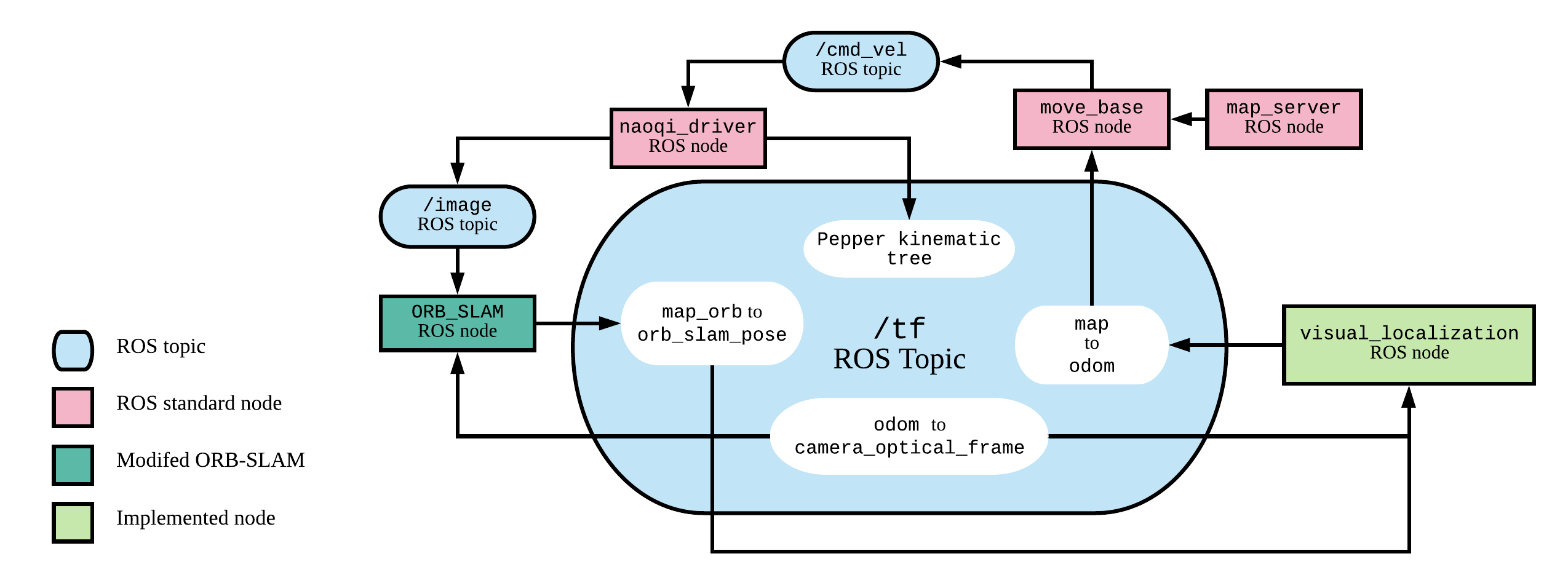}
	\caption{Overview of our proposed system. The camera images are feed to the ORB-SLAM system together with the camera position with respect to the odometry frame (\emph{odom}).  An estimated camera position with respect to an arbitrary fixed frame is given as output by ORB-SLAM. The \texttt{visual\_localization} node takes this information and the Pepper kinematic information to compute a transformation between the standard fixed frame \emph{map} and the \emph{odom} frame.}
	\label{fig:pipeline}
\end{figure}

%Our visual navigation system for Pepper consists of two main steps: First, the localization step is based on the visual SLAM system aided by wheel odometry, which provides not only localization with respect to a given frame but also a map representation. The second step uses the map and the current localization estimate to and projects them into the 2-dimensional space to perform planar planning.
Our visual SLAM-based localization and navigation system for Pepper consist of three main components, which are shown in Figure \ref{fig:pipeline}. Firstly, an ORB-SLAM-based localization and mapping system, which uses a single RGB camera located in Pepper's forehead, and odometry measurements computed by the proprietary Pepper's software. The second component correspond to the \texttt{visual\_localization} \footnote{\url{https://github.com/ristofer/visual_localization/}} ROS node that transforms ORB-SLAM's camera pose estimate to a transformation between the standard \emph{map} frame and the \emph{odom} frame. Finally, the node \texttt{move\_base} \footnote{\url{http://wiki.ros.org/move_base}} executes the navigation process. 
%the mapping process using the Visual-Inertial SLAM system.  The second step is using only the localization system that ORB-SLAM provides. This only-localization system is much more lightweight than the full SLAM system.  Finally, the position information provided by the localization system is given to the ROS Navigation Stack in a standard format.  The Pepper's lasers are used to generate a cost map for the Navigation Stack.

\subsection{ORB-SLAM-based localization}
Our localization system maintains the same software architecture with 3 parallel threads, original from ORB-SLAM2 \cite{Mur-Artal2017}: incoming images are processed in the \emph{Tracking} thread, creating new map points and estimating the current camera pose $ \notationMatrix{T}{CW}$ in world frame $\notatFrame{W}$; a \emph{Local Mapping} thread which builds on the map and the keyframes and frequently performs local bundle adjustments to update the positions of map points and camera poses at the keyframes; a \emph{Loop Closing} thread which detects loops in the trajectory and propagates a correction through the trajectory poses and the map. In addition, we implemented the following improvements:

% \begin{figure}[h]
% 	\centering
% 	\missingfigure{A diagram of the ORB-SLAM System would be good}
% 	\caption{ORB-SLAM architecture.}
% 	\label{fig:reference-frames}
% \end{figure}

\subsubsection{Tracking modifications}
We changed the Tracking thread to process not only images but also odometry measurements, obtained directly from ROS. Odometry measurements are computed within the Pepper's internal software and published in ROS through NAOqi wrappers with respect to the \emph{odom} frame, shown as $\notatFrame{O}$ in Figure \ref{fig:reference-frames}. Our ROS-compatible wrapper for ORB-SLAM subscribes the \texttt{tf} topic and images, and requests an odometry measurement every time a new image is received, obtaining a synchronized pair image-odometry. Later, every time a new keyframe is created after a successful camera tracking, the odometry information is also included in the keyframe.

In addition, since the original behavior of ORB-SLAM  is to stop providing camera poses when camera tracking fails, and wait until a relocalization, which is not a desirable strategy while navigating \footnote{Unless high-level behaviors to detect failures are considered}, we set the camera estimation equal to the odometry prediction. This ensures a continuous camera pose hypothesis for planning tasks but requires that the metric scale is initialized.

\subsubsection{Metric scale initialization}
We did not utilize any general system initialization solution as in \cite{Martinelli2014} but preferred a multi-step approach as in \cite{Mur-Artal2016a}. We first wait until the pure visual SLAM system is initialized and the unscaled map built, to then compute the scale from the odometry information between keyframes.
%The odometry obtained from the 2D base model of Pepper provides a complementary addition to ORB-SLAM, considering that the odometry provides strong information on the self-motion of the robot and, amongst others,the map scale ambiguity can be resolved. 
By comparing the relative translations between keyframes as predicted by ORB-SLAM $\Delta \notationVector{p}{O}(i-1,i)$ and the odometry $\Delta \notationVector{p}{W}(i-1,i)$, the scale can be retrieved and the map and keyframe poses can be updated by the method of Horn \cite{Horn1987} (Eq. \eqref{eq:escale-1}). However, the initial map is subject to major change in the early stages of the mapping. Therefor the scale correction is done after a fixed number of \emph{N} keyframes have been created, ensuring a satisfactory converged map and thus a reasonably reliable scale correction. The success of this strategy only depends on the environment's size and the motion performed by the robot; an additional discussion is given in Section \ref{sec:results}.
\begin{equation}
\label{eq:escale-1}
\notatScalar{s} = \frac{\sqrt{\sum_{i}^{N} \lVert \Delta \notationVector{p}{O}(i-1,i)\rVert^{2}}}{\sqrt{\sum_{i}^{N} \lVert \Delta \notationVector{p}{W}(i-1,i) \rVert^{2}}}
\end{equation}

%The number of keyframes depends of the environment's size. A large environments needs more keyframes to correctly calculate the scale.\\
% The relative pose of the camera between time $k$ and $k + 1$ as predicted by both ORB SLAM and the odometry can be obtained as follows:
% \begin{equation}
% \notationMatrixFrame{T}{O}{C} = \notationMatrixFrame{T}{O}{C_{k}W} \cdot \notationMatrixFrame{T}{O}{WC_{k+1}} .
% \end{equation}
% This operation cancels out any accumulated error in the odometry by taking only a small temporal difference of the total movement. 
% The scale can be obtained by solving the linear equation  $A_{n \times 3}   \textbf{s}  = B_{n \times 3}$ through a Singular Value Decomposition, between the relative translations of these poses given by ORB-SLAM and the odometry, where n is the number of keyframes used for the scale recovery. 

After the scale update, a Global Bundle Adjustment (Global BA) is performed to guarantee an optimal map reconstruction. 

\subsubsection{Local Mapping}
Every time a new keyframe is created, \emph{Local Mapping} performs an optimization in a subset of the complete trajectory updating both the poses and the map -the so-called \emph{local window}. The parts of the trajectory to be optimized are keyframes in the neighborhood of the last added keyframe, and also map points being observed by those; the neighbors are selected by the so-called \emph{covisibility graph}\cite{Mur-Artal2015a}. This operation on the local window ensures an efficient optimization process even in large maps.

Since the initialization procedure makes the current trajectory and map \emph{metrically consistent}, it is possible to fuse the visual information with wheel odometry information to avoid drift. This is done by adding odometry factors or \emph{constraints} between keyframes. In order to do so, the odometry measurement is mapped from the odometry frame $\notatFrame{O}$ to the camera frame $\notatFrame{C}$ by using Pepper's forward kinematics. Hence, we compute the difference between the odometry measurements $i-1$ and $i$, $\notationMatrixFrame{T}{C}{C_{i-1}, C_{i}}$, between all the keyframes in the local window, which hopefully match the difference between the keyframes' pose, $(\notationMatrix{T}{C_{i-1} W}  \notationMatrix{T}{C_{i} W})$. The error between the odometry's and ORB-SLAM's differences are defined in the minimal representation of the pose, i.e. 6-dimensional, which is achieved by using the \emph{logarithm map} of $\SEthree$:
\begin{equation}
\label{eq:odo-res}
\varepsilon_{odo} = \LogmapSEthree{\notationMatrixFrame{T}{C}{C_{i-1}, C_{i}}^{-1}  \notationMatrix{T}{C_{i-1} W}  \notationMatrix{T}{C_{i} W}^{-1} } .
\end{equation}

This residual is defined for every pair of keyframes within the local window; additionally, keyframes with neighbors which are not in the local window, are also added as fixed nodes in the optimization. The corresponding optimization problem that minimizes visual error terms $\varepsilon_{vis}$ (as defined in \cite{Mur-Artal2015a}) and odometry terms $\varepsilon_{odo}$ (Eq. \eqref{eq:odo-res}), is:
\begin{equation} \label{eq:local-ba}
\notatManifold{\optimum{X}} = \argmin_\notatManifold{X} = {\displaystyle \sum_{(i,k)} \mahalanobisNorm{ \varepsilon_{vis} }{\notationMatrix{\Omega}{vis}}  + \sum_{(i-1,i)} \mahalanobisNorm{ \varepsilon_{odo} }{\notationMatrix{\Omega}{odo}}}
\end{equation}

The optimization problem in equation \eqref{eq:local-ba} is solved with the graph optimization framework \emph{g2o}\cite{Kummerle2011a} using fixed information matrices $\notationMatrix{\Omega}{odo}, \notationMatrix{\Omega}{vis}$. The resulting keyframe poses and map points are then updated, and the \emph{Local Mapping} thread awaits until a new keyframe is added from \emph{Tracking}.

\subsubsection{Localization mode and map reuse}
ORB-SLAM provides the option to localize in a previously built map, disabling the SLAM capabilities.  This localization can run in a single thread, hence requiring a fraction of the computational requirements compared to the full ORB-SLAM system. Nevertheless, in order to perform localization-only, it is required a map that was built in the same session.

Since this is not generally the case, we use map saving capabilities (taken from a fork of ORB-SLAM \footnote{\url{https://github.com/Alkaid-Benetnash/ORB_SLAM2}}) and implemented a different behavior for the system when it is launch with a pre-built map, that first tries to relocalize and then continue mapping incrementally. These minor changes allowed us to build maps even during different sessions once the relocalization is successful.
\subsection{Navigation}
%The previous section presented our odometry-aided visual localization system, which is able to build maps incrementally in a SLAM fashion. However, for the majority of RoboCup @Home tests, the SLAM capabilities are not of interest since the environments are known in advance. 
For the navigation part, we assume the ORB-SLAM's map was already built, so we can rely on the localization mode.  
%For RoboCup @Home tests where an actual SLAM system is needed \cite{Iocchi2015}, e.g. \emph{Help Me Carry} , \emph{Supermarket}, \emph{Walk \& Talk}  and the old \emph{Restaurant}, the SLAM mode can be used, always considering that SLAM mode is more computationally expensive than localization mode.
%Navigation in wheeled robots has the advantage of being less computationally demanding that other platforms since the displacement is planar. It is for this reason that LIDARs are the standard to approach the problem. 
The pose estimation performed by ORB-SLAM is 6-dimensional since it considers a camera freely moving in the space, which would be an overkill to perform planning with Pepper. In order to use ORB-SLAM's estimates within a planar navigation framework, we developed the \texttt{visual\_localization} node, which takes the estimated position of the camera with respect to the ORB-SLAM world frame and computes a transformation between the ROS standard \emph{map} and \emph{odom} frames. This transformation represents the Pepper position in the ORB-SLAM map based on the estimated pose of the Pepper's camera and its kinematic information.

The \texttt{move\_base} package is used to navigate.  Our localization system basically replaces the \texttt{amcl} \footnote{\url{http://wiki.ros.org/amcl}} package in the ROS Navigation Stack. The \texttt{move\_base} package uses the pose estimate provided by the localization system and Pepper's laser readings to compute the cost map necessary for planning. Thus, lasers are not used for localization, but for obstacle detection and path planning.

% It uses the position of the robot based on the information given by our localization system is used by this package. Normally the \texttt{amcl} \footnote{\url{http://wiki.ros.org/amcl}} system is used for the localization.  Our system only replaces the \texttt{amcl} system.

%	The localization system provides a transformation between a fixed frame, defined by the initialization of ORB-SLAM, and the estimated camera optical frame.  This information is used in conjunction with the kinematic information of Pepper to calculate a transformation between a new frame called map, that correspond to the same fixed frame defined by ORB-SLAM, and the odom frame of the robot.  This way, the robot's position is given by knowing the camera's position.

%With the map to odom transformation and odom to base link transformation is possible to use the ROS Navigation Stack.  To  detect obstacles, a laser scan is needed.  We use the Pepper's laser for this task.

%%%%%%%%%%%%%%%%%%%%%%%%%%%%%%%%%%%%%%%%%%%%%%%%%%%%%%%%%%%%%%%%%%%%%%%%%%%%%%%%%%%%%%%%%%%%%%%%%%%%%%%%%%%%
\section{Experiments and results}
\label{sec:results}

\subsection{Experimental setup}
%\subsubsection{Places}
We considered two real environments of the Faculty of Physical and Mathematical Sciences of Universidad de Chile to test our system: \emph{Mechatronics Laboratory} and \emph{School Building South Hall}. The chosen places were different in size, furniture, and visual features complexity, being the latter of paramount importance for the visual SLAM system.

\begin{figure}[H]
	\centering
	\includegraphics[width=0.31\textwidth]{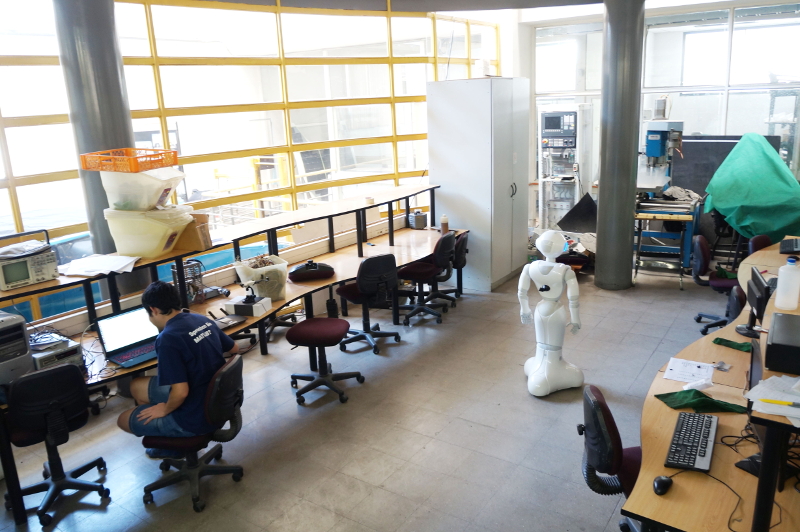}
    \includegraphics[width=0.31\textwidth]{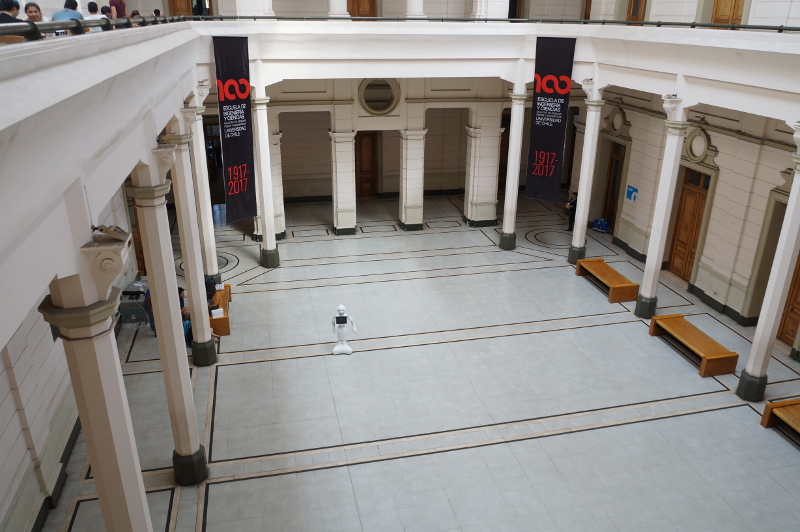}
	%\caption{\emph{Left:} Mechatronics Lab. \emph{Center:} Chemistry Building Hall. \emph{Right:} South Hall}
    \includegraphics[width=0.31\textwidth]{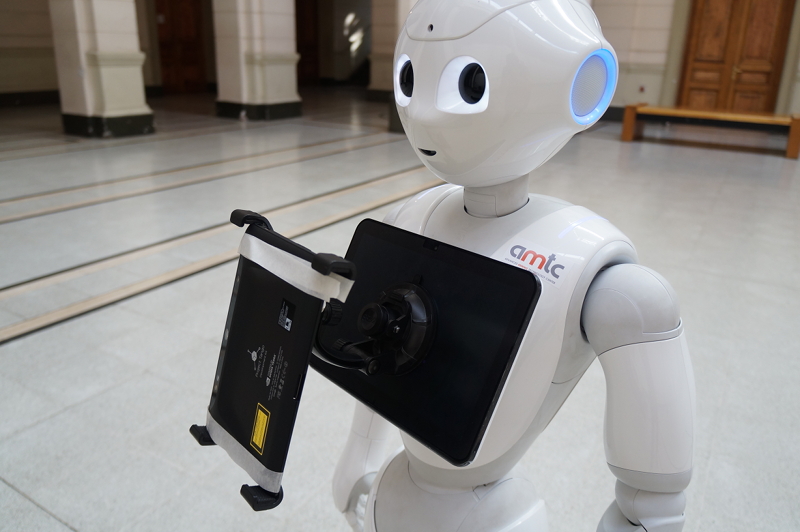}
    \caption{\emph{Left:} Mechatronics Lab. \emph{Center:} South Hall. \emph{Right:} Pepper with Google Tango Tablet attached for ground truth measurements.}
	\label{fig:environments}
\end{figure}

\begin{itemize}
\item The Mechatronics Laboratory (Figure \ref{fig:environments}a)  is a 10x9$m^2$ space. The main furniture are rolling chairs and work tables. It is a feature-rich space comparable to the RoboCup Arena; however, it has various windows that enable the pass of natural light.  
%\item The Chemistry Building Hall (Figure \ref{fig:environments}b) is an open space of 12x17$m^2$. It has four benches in the middle and some posters.  The main light sources are a skylight and artificial light. This place is feature-rich but they are mainly on the walls.
\item The School Building South Hall (Figure \ref{fig:environments}b) has an area of 16x27.5$m^2$. It is an open space with pillars and doors, but generally feature-less, making it the most challenging environment for our system.
\end{itemize}

To have a ground truth reference, a Google Tango Tablet is used (Figure \ref{fig:environments}c).

%\subsubsection{Evaluation platforms}
%Our experiments were developed with the Pepper robot connected to an external standard laptop (i7, 8Gb RAM) through a WiFi connection provided by a 5GHz router. The laptop was only running ORB-SLAM. The ROS/NaoQi driver was run on the Pepper computer. In addition, we used a \emph{Google Tango Tablet Development Kit}\footnote{Currently unsupported by Google.} as ground truth system because it is an \emph{engineered} integrated visual-inertial solution.  The Google Tango Tablet was attached to Pepper through a suction cup tablet stand.

\subsection{Experiments}
\subsubsection{Mapping}
%\begin{figure}[H]
%	\centering
%\includegraphics[width=0.8\textwidth]{quimica-mapping.png}
%	\caption{Mapping test on Chemistry Building Hall. The scale does not get correctly initialized.}
%	\label{fig:chemistry-mapping}
%\end{figure}
The first experiment considered a localization and mapping task; this was performed in both the Mechatronics Lab (Fig. \ref{fig:environments}a) and South Hall (Fig. \ref{fig:environments}b). We remote controlled the robot to build a three dimensional map to be later used for localization.  Table \ref{table:ate} compare different mapping results through the \textit{Absolute Trajectory Error} \cite{Sturm2012}, a metric that calculates the root mean square error $RMSE$ defined as  $\left( \frac{1}{N} \sum_{i}^{N} \left \| p_{e_i} - p_{gt_i} \right \|^2 \right) ^{1/2}$ between the localization estimate $p_{e_i}$ and the ground truth $p_{gt_i}$ through all the time indices.

\setlength{\tabcolsep}{9pt}
\begin{table}[]
  \centering
  \caption{Absolute Trajectory Error (ATE) in meters, for each place and axis. A mapping experiment was performed in the Mechatronics Laboratory and in the South Hall. The estimated trajectory and the ground truth was used to calculate the ATE.}
  \label{table:ate}
  \begin{tabular}{lccc}
  Place                    & ATE X [m]          & ATE Y [m]         & ATE Z [m]            
  \\ \hline
  Mechatronics Laboratory   & 0.270 & 0.249 & 0.080   
  \\
  %Chemistry Building Hall  & --             & --             & --                
  %\\
  South Hall                & 0.619 & 0.849 & 0.390  
  \end{tabular}
\end{table}

During all the experiments we noticed that the robot must move smoothly and preferably sideways in order to triangulate the initial map; pure rotational factors must be avoided despite the offset between the head camera and the base's axis of rotation. The initial displacement is primordial to recover a reliable scale factor as well. However, this also depends on a parameter that sets the number of keyframes to wait until the scale is recovered with Eq. \ref{eq:escale-1}, which is set empirically. 

Regarding mapping, as is expected from a feature-based visual SLAM system, the number of points and quality of the map increases in feature-rich environments. In addition, compared to LIDAR mapping, visual mapping requires significantly more time. This because map creation depends on the field-of-view (FOV) of the camera, which is very narrow in Pepper, requiring to map the same place from multiple views in order make it useful for robust localization. LIDAR does not suffer from this issue since localization is performed by point cloud alignment rather than feature matching. However, feature matching has the advantage of providing instantaneous relocalization when the robot is lost since places are uniquely defined by a bag-of-words representation \cite{Mur-Artal2015a}.

\subsubsection{Localization and Navigation}
We performed a second experiment to test the localization and navigation in a known place, i.e., with a pre-built map. This was also executed in the Mechatronics Laboratory and South Hall.

\begin{figure}[h]
	\vspace{-0.5cm}
    \centering
\includegraphics[width=0.8\textwidth]{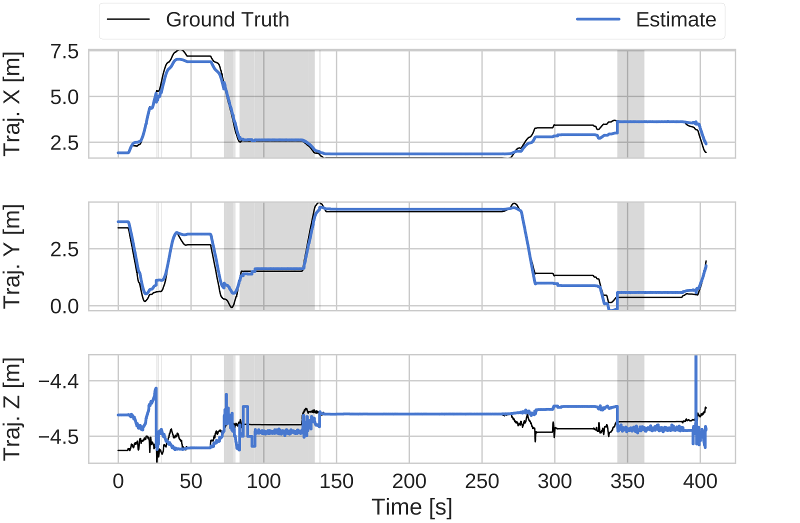}
	\caption{Navigation test on the Mechatronics Laboratory.  The estimate of the localization system is compared to ground truth. Grey areas in the graph indicated when the robot is correctly localized with ORB-SLAM. When the robot is not localized, an odometry estimate is used}
	\label{fig:mechatronics-nav}
\end{figure}

\begin{figure}[h]
	\centering
\includegraphics[width=0.8\textwidth]{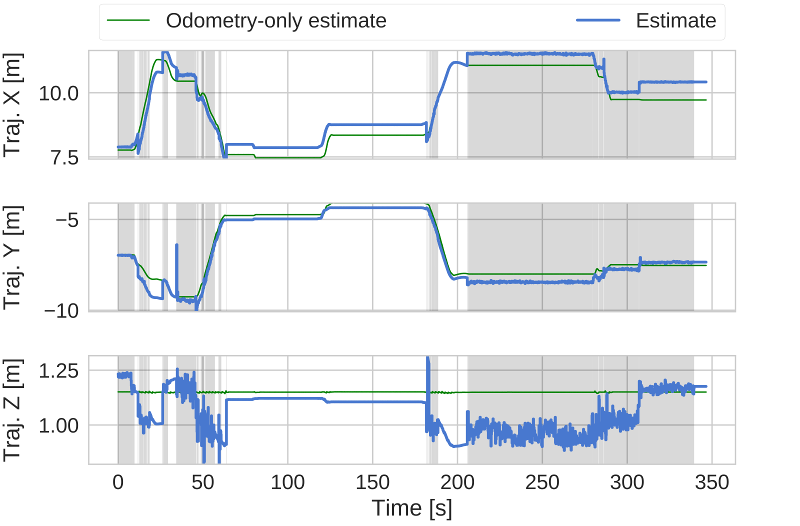}
	\vspace{-0.2cm}
	\caption{Navigation test on the South Hall. The robot tries to navigate but the localization system does not work correctly.}
	\label{fig:south-hall-nav}
    \vspace{-0.7cm}
\end{figure}

We commanded the robot to navigate without operator help to a relative point with respect to its initial pose, which exploited the localization capabilities of our system in a known environment.  Localization results are shown in Figures \ref{fig:mechatronics-nav} and \ref{fig:south-hall-nav}.

Our experiments show the performance of the system, which uses both visual localization and odometry fusion (highlighted in gray) and odometry-only localization when the robot is lost (in white). In the navigation experiment in the Mechatronics Lab, showed in Figure \ref{fig:mechatronics-nav}, Pepper correctly navigated through the test. However, between seconds 275 and 350 there exists a considerable drift between the ground truth and the localization estimate. These problems can result in reaching an erroneous navigation goal or even collide if no safety procedures are considered. We believe that a cause of this issue was the lack of viewpoints during the mapping step, as mentioned in the previous experiment.

%This can translate into a failed navigation goal or even a collision. A solution to this is to create a better map with images captured from various viewpoints to increment the overall localization time.

Regarding South Hall experiments, the multiple discontinuities in the localization estimate (Figure \ref{fig:south-hall-nav}, Z axis) made navigation unfeasible. This was caused by the large distance between the robot and the landmarks in this environment, which was not the case of the Mechatronics Lab. Since visual SLAM systems are based on optimization and Pepper's FOV is narrow, it is more difficult to correctly estimate the pose because the triangulation uncertainty is higher; this is a known problem in visual systems \cite{Hartley2004}.

%The distance between the camera and the features are larger in the South Hall than in the Mechatronics Laboratory.  This situation creates ambiguities in the viewpoint calculation that translate into an inaccurate pose estimate.  Moreover when the system re-localize the pose estimate can perform a jump.

However, in both experiments we noticed that localization is robust to changes in the environment, like a change in furniture position and, if the map is correctly built, there is minimal (if not zero) accumulation of drift error. %, but errors in scale calculation can be a problem when performing large translations. %The localization works better with images with plenty of features.  In the future, it is ideal to implement a system that rates the viewpoints based on the number of features and the localization probability.

%To keep localized while moving, the movements have to be smooth.  The typical translation move of Pepper is smooth enough for the localization process.  Localization is robust to occlusion. Great changes in illumination negatively affect the localization process. The Figure \ref{fig:mechatronics-nav} shows that the overall time that the robot is localized is near 20\%.  This is mainly because the environment had different illumination when the map was built and when the navigation experiment was done.  Also, some objects where moved (the navigation test was performed a week after the mapping process).  However, even when the robot is not always localized through ORB-SLAM, the odometry estimation is enough for short linear translations.

\subsection{Discussion}
\label{subsubsec:discussion}
The previous experiments evidence several advantages but also challenges of our proposed system. We summarizes them as follows: 
\paragraph{Advantages} Our localization system is not affected by Peppers' LIDARs short range, which is one of the main limitations of it in RoboCup environments. Since we used the map created visually, the robot is able to localize with a single look by exploiting the relocalization capabilities of ORB-SLAM. Our scale recovery solution also allowed us to perform metric mapping despite using a single camera. In addition, since the motion estimation is based on features, it is robust to partial occlusions, and odometry is used when no visual features are tracked. All these advantages demonstrate that it is possible for our system to localize the robot in RoboCup@Home arenas successfully.
\paragraph{Challenges} Despite the previous advantages, we cannot avoid to mention some challenges and difficulties we noticed during our experiments. The first one relates to illumination changes, which can deteriorate hugely visual tracking. If we set the camera exposure to automatic we can deal with dynamic lighting, but the system is more susceptible to motion blur, which is still an issue despite the robot performs planar motion; the main cause of this is joint backslash. If the environment has non-variable illumination, we recommend to fix the camera exposure to diminish those problems. The second challenge regards glossy surfaces, which produce fake landmarks because of reflections. Despite ORB-SLAM is able to deal with outliers that do not match the predicted motion, it is still an open challenge in our opinion. Finally, localization turns noisier when the landmarks are far away, which is caused by the optimization procedures and the point's triangulation uncertainty.
\section{Conclusions}
\label{sec:conclusions}

In this work, we presented a localization system for a Pepper robot based on a visual SLAM system. Our solution, built upon ORB-SLAM, was focused on developing a self-localization system able to deal with large environments despite the LIDARs' short range. In order to do so, we presented an approach that fused visual and wheel odometry information. We tested the system in two real environments, showing the feasibility performing SLAM and navigation with our system with the current Pepper sensors, despite displaying some issues such as weakness to illumination changes, ambiguities to glassy surfaces and far landmarks. 

Nowadays we are working towards an on-board implementation of the self-localization system on Pepper, which will allow us to perform a more exhaustive evaluation and comparison with other sensors such as lasers. In the future, we would like to improve robustness to illumination changes and reducing the noisy behavior in large environments.

\section*{Acknowledgments}
\label{sec:ack}
This work was partially funded by the FONDECYT 1161500 project.

%%%%%%%%%%%%%%%%%%%%%%%%%%%%%%%%%%%%%%%%%%%%%%%%%%%%%%%%%%%%%%%%%%%%%%%%%%%%%%%%%%%%%%%%%%%%%%%%%%%%%%%%%%%%
% Bibliography
\bibliography{orbib}
\end{document}

%% file: packages.tex
\include{definitions}

\usepackage{fancyhdr}

\usepackage[latin9]{inputenc}

\usepackage{psfrag}%needed to replace text in figures with latex text
\usepackage{amssymb,amsfonts}%for equations
\usepackage{units} %to use the correct spacing between number and unit

\usepackage{dsfont,latexsym,cite, comment, graphicx,color}

\usepackage{subfigure}

\usepackage{todonotes}	% to include 'to do' messages in the paper

\usepackage[hyphens]{url}
\usepackage[pdfpagelabels]{hyperref}

\usepackage{xcolor}

\usepackage{float}

%% file: definitions.tex
\usepackage{amsmath}
\usepackage{amssymb}
\usepackage{amsfonts}
\usepackage{dsfont}
\usepackage{amssymb}% http://ctan.org/pkg/amssymb
\usepackage{pifont}% http://ctan.org/pkg/pifont
\usepackage{etoolbox}	% robustify command

\newcommand{\etal}{\emph{et al.}\ }

\newcommand{\mahalanobisNorm}[2]{\lVert{#1}\rVert^{2}_{#2}}

\newcommand{\optimum}[1]{{#1}^{*}}

\newcommand{\argmin}{\operatornamewithlimits{arg\,min}}
\newcommand{\subarrow}[1]{
	\mathord{
		\renewcommand{\arraystretch}{0}
		\begin{array}[t]{@{}c@{}l@{}}
			#1\\[2pt]
			\hspace{-2pt}\scriptstyle\longrightarrow
		\end{array}
		\kern\scriptspace
	}
}
\newcommand{\notatFrame}[1]{\subarrow{\mathcal{F}}{}_{\scriptscriptstyle #1}}

\newcommand{\notatMatrix}[1]{\boldsymbol{\mathrm{#1}}}
\newcommand{\notatVector}[1]{\boldsymbol{\mathrm{#1}}}
\newcommand{\notatScalar}[1]{{#1}}
\newcommand{\notatHomog}[1]{\boldsymbol{{#1}}}
\newcommand{\notatManifold}[1]{\mathcal{\MakeUppercase{#1}}}

\newcommand{\notationMatrix}[2]{\boldsymbol{\mathrm{#1}}_{\scriptscriptstyle #2}}
\newcommand{\notationVector}[2]{\boldsymbol{\mathrm{#1}}_{\scriptscriptstyle #2}}
\newcommand{\notationScalar}[2]{{#1}_{\scriptscriptstyle #2}}
\newcommand{\notationHomog}[2]{\boldsymbol{{#1}}_{\scriptscriptstyle #2}}
\newcommand{\notationManifold}[2]{\mathcal{\MakeUppercase{#1}}_{\scriptscriptstyle #2}}

\newcommand{\notationMatrixFrame}[3]{{\scriptscriptstyle_#2}\boldsymbol{\mathrm{#1}}_{\scriptscriptstyle #3}}
\newcommand{\notationVectorFrame}[3]{{\scriptscriptstyle_#2} \boldsymbol{ \mathrm{#1}}_{\scriptscriptstyle #3}}
\newcommand{\notationScalarFrame}[3]{{\scriptscriptstyle_#2}{#1}_{\scriptscriptstyle #3}}
\newcommand{\notationHomogFrame}[3]{{\scriptscriptstyle_#2}\boldsymbol{{#1}}_{\scriptscriptstyle #3}}

\robustify{\notatFrame}
\robustify{\notatMatrix}
\robustify{\notatVector}
\robustify{\notatScalar}
\robustify{\notatHomog}
\robustify{\notationMatrix}
\robustify{\notationVector}
\robustify{\notationScalar}
\robustify{\notationHomog}
\robustify{\notationMatrixFrame}
\robustify{\notationVectorFrame}
\robustify{\notationScalarFrame}
\robustify{\notationHomogFrame}

\newcommand{\SOthree}{\mathrm{SO(3)}}

\newcommand{\SEthree}{\mathrm{SE(3)}}

\newcommand{\LogmapSEthree}[1]{\mathrm{Log}_{\SEthree}\left(#1\right)}

\makeatletter
\renewcommand*\env@matrix[1][\arraystretch]{%
	\edef\arraystretch{#1}%
	\hskip -\arraycolsep
	\let\@ifnextchar\new@ifnextchar
	\array{*\c@MaxMatrixCols c}}
\makeatother

\usepackage{color}
\usepackage{colortbl}
\definecolor{ColorLightCyan}{rgb}{0.88,1,1}
\definecolor{ColorLightTurquoise}{rgb}{0.5, 1, 0.8}